%% file: root.tex
\documentclass[letterpaper, 10 pt, journal, twoside]{IEEEtran}  

\IEEEoverridecommandlockouts                              
\usepackage{graphicx} 
\usepackage{amsmath} 
\usepackage{amssymb}  
\usepackage[usenames, dvipsnames]{color}
\usepackage{siunitx}
\usepackage{cases}
\usepackage{lipsum}
\usepackage{color}
\usepackage{cite}
\usepackage[ruled,vlined]{algorithm2e}
\usepackage{diagbox} 
\usepackage{tabu}
\usepackage{array}
\usepackage{multirow}
\usepackage{booktabs}
\usepackage{makecell}
\usepackage{empheq}
\usepackage{threeparttable}
\usepackage{tabularx}
\usepackage{hyperref}

\newcolumntype{P}[1]{>{\centering\arraybackslash}p{#1}}
\newcolumntype{M}[1]{>{\centering\arraybackslash}m{#1}}

\newcommand{\revision}[1]{\textcolor{black}{#1}} 
\newcommand{\final}[1]{\textcolor{black}{#1}} 
\newcolumntype{C}[1]{>{\centering\let\newline\\\arraybackslash\hspace{0pt}}m{#1}}
\newcolumntype{L}[1]{>{\raggedright\let\newline\\\arraybackslash\hspace{0pt}}m{#1}}

\title{
Gaze-Guided Robotic Vascular Ultrasound Leveraging Human Intention Estimation
}

\author{Yuan Bi$^{1, 2}$, Yang Su$^{1}$, Nassir Navab$^{1, 2}$, \textit{Fellow, IEEE}, and Zhongliang Jiang$^{1, 2}$ 

\thanks{Manuscript received: September 10, 2024; Revised: December 7, 2024; Accepted: January 15, 2025.}
\thanks{This paper was recommended for publication by Editor Jessica Burgner-Kahrs upon evaluation of the Associate Editor and Reviewers' comments.}
\thanks{This work was partially supported by SINO-German Mobility Project (M-0221). (Corresponding Author: Zhongliang Jiang, zl.jiang@tum.de)}
\thanks{$^{1}$Y. Bi, Y. Su, N. Navab, and Z. Jiang are with the Chair for Computer-Aided Medical Procedures and Augmented Reality, Technical University of Munich, Boltzmannstr. 3, 85748 Garching, Germany}
\thanks{$^{2}$Y. Bi, N. Navab, and Z. Jiang are with the Munich Center for Machine Learning, Arcisstraße 21, 80333 M\"unchen, Germany}
\thanks{Digital Object Identifier (DOI): see top of this page.}
}

\begin{document}

\maketitle

\begin{abstract}
Medical ultrasound has been widely used to examine vascular structure in modern clinical practice. However, traditional ultrasound examination often faces challenges related to inter- and intra-operator variation. The robotic ultrasound system (RUSS) appears as a potential solution for such challenges because of its superiority in stability and reproducibility.
Given the complex anatomy of human vasculature, multiple vessels often appear in ultrasound images, or a single vessel bifurcates into branches, complicating the examination process.
To tackle this challenge, this work presents a gaze-guided RUSS for vascular applications. A gaze tracker captures the eye movements of the operator. The extracted gaze signal guides the RUSS to follow the correct vessel when it bifurcates. Additionally, a gaze-guided segmentation network is proposed to enhance segmentation robustness by exploiting gaze information.
However, gaze signals are often noisy, requiring interpretation to accurately discern the operator's true intentions. To this end, this study proposes a stabilization module to process raw gaze data. The inferred attention heatmap is utilized as a region proposal to aid segmentation and serve as a trigger signal when the operator needs to adjust the scanning target, such as when a bifurcation appears. To ensure appropriate contact between the probe and surface during scanning, an automatic ultrasound confidence-based orientation correction method is developed.
In experiments, we demonstrated the efficiency of the proposed gaze-guided segmentation pipeline by comparing it with other methods. Besides, the performance of the proposed gaze-guided RUSS was also validated as a whole on a realistic arm phantom with an uneven surface.
\end{abstract}



\bstctlcite{IEEEexample:BSTcontrol}
\input{text}


\bibliographystyle{IEEEtran}
\bibliography{IEEEabrv,references}

\end{document}

%% file: text.tex
 \section{Introduction}
Medical ultrasound, valued for its non-invasiveness, portability, real-time performance, and affordability, is widely used in clinical practice for screening and intra-operative guidance. However, traditional free-hand ultrasound suffers from inter- and intra-operator variances. Image quality depends on acquisition parameters like contact forces, angles, and probe positioning~\cite{tan2023autonomous,huang2018robotic}, making it highly operator-dependent and reducing result reproducibility~\cite{tan2022flexible}. To tackle such a dilemma, robotic ultrasound systems (RUSS) offer a promising solution to address these challenges~\cite{jiang2023robotic, akbari2021robot,bi2024machine}.

\par
Owing to the fast development in the field of robotics, robots have shown their superiority over humans in terms of stability and precision.
Working together with human experts as colleagues, the interactive RUSS can largely release the burden of cumbersome acquisition from sonographers and greatly enhance the reproducibility of the imaging process~\cite{li2021overview,von2021medical,huang2023review}.
RUSSs are used for autonomous vascular ultrasound screening~\cite{jiang2021autonomous,huangQ2024robot}, but the complexity of vessel distributions and branches often results in multiple vessels or bifurcations in images. Explicit human guidance is essential for online robotic adjustments to better visualize target vessels. This guidance also aids segmentation networks as region proposals, improving vessel segmentation, particularly for small limb vessels ($2.70\pm0.15$ mm)~\cite{wahood2022radial}, which are challenging to extract accurately from B-mode images.

\par
In order to tackle the aforementioned challenges, this work proposes a gaze-guided RUSS. Gaze tracker, as an intuitive and efficient tool for human-machine interaction, has been widely applied to various medical robotic systems~\cite{noonan2008gaze,tong2015retrofit,guo2019novel}. This study presents the first exploration of integrating gaze tracking into RUSSs, allowing the system to incorporate real-time human supervision to improve the intelligence and robustness of the RUSS during complex scanning tasks. 
Especially in intra-operative ultrasound imaging, the hands of surgeons are often occupied with essential surgical tools such as needles or catheters, making manual maneuvering of the ultrasound probe challenging. Implementing a gaze tracker as a part of the RUSS helps the operators control the ultrasound scanning without disrupting the surgical workflow.
On one hand, the eye tracking information can indicate the vessel of interest during the scanning. On the other hand, the gaze of the operator can also be applied as network attention to assist the vessel segmentation, particularly for challenging ones with limited size. Based on the segmentation results, the robot is controlled to adapt its maneuvers so that the vessel of interest is kept in the middle of the ultrasound image. To further improve the ultrasound image quality during scanning, a confidence-driven control for linear ultrasound probes is proposed. The main contributions of this article are listed as follows:\footnote{\final{Codes: \url{https://github.com/yuan-12138/Gaze_RUSS}}}\footnote{\final{Video: \url{https://www.youtube.com/watch?v=Zul2-fqy0FE}}}
\begin{itemize}
  \item \revision{A gaze-guided segmentation network is proposed, which effectively integrates attention heatmaps derived from gaze patterns into segmentation. This improves segmentation performance by enabling the system to focus on the relevant regions in ultrasound images and providing the ability to segment vessels based on the operator's gaze.}
  \item  \revision{A human intention estimation module is introduced to predict the operator's intention during scanning by combining gaze patterns with segmentation result history. This module stabilizes noisy gaze signals, providing consistent and reliable guidance to the segmentation network.}
  \item \revision{A gaze-guided RUSS is developed and validated on a realistic arm phantom, showcasing its ability to perform complex scans. The system dynamically switched focus between vessels based on the operator's eye tracking, while the confidence-driven control method continuously adjusted probe orientation to optimize image quality.}
\end{itemize}

\par
To emphasize realism, we highlight that the proposed gaze-guided RUSS does not rely on highly stable gaze input from clinicians. The system can still operate with the intermittent absence of gaze input by using a zero-value map as the gaze attention heatmap (see Sec.~\ref{sec_gaze_guided_seg}). When necessary it can seamlessly integrate the operator's input, such as when adjusting the scanning focus for complex scenarios like bifurcations.

\begin{figure*}[ht!]
\centering
\includegraphics[width=0.8\textwidth]{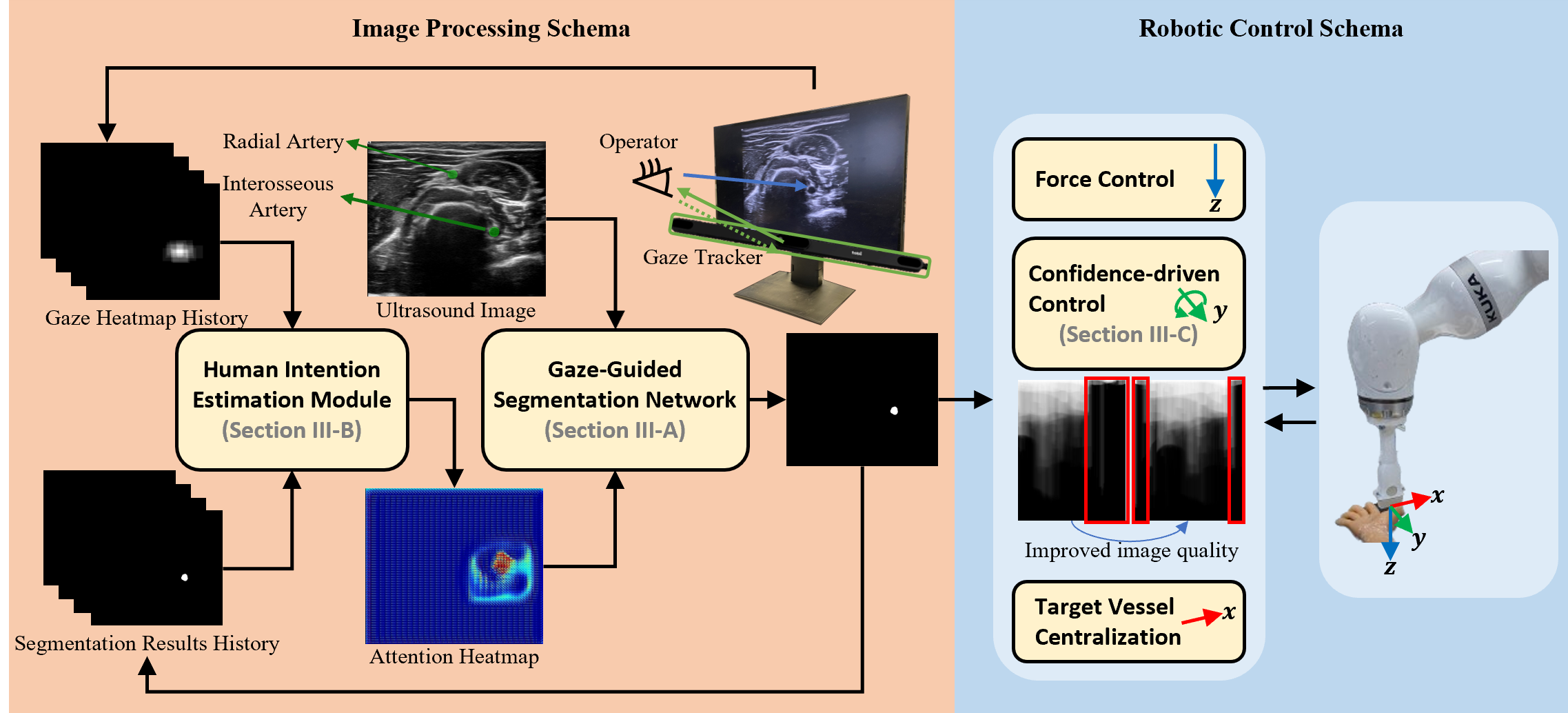}
\caption{
Overview of the Proposed Gaze-Guided Interactive RUSS: The human gaze signal, captured by a gaze tracker, is combined with segmentation history in an intention estimation module to infer the operator's preference, especially when multiple vessels are visible. The resulting attention heatmap guides the gaze-guided segmentation network to produce accurate vessel segmentation masks. These results are integrated into the robotic control loop to keep the target vessel centered in the ultrasound image. A confidence-based orientation correction optimizes probe contact with curved surfaces, improving image quality. Red boxes in ultrasound images highlight shadowed areas caused by improper probe contact.
}
\label{Fig_overview}
\end{figure*}

\section{Related Work}
\subsection{Robotic Ultrasound Systems}
Ultrasound image quality depends heavily on acquisition parameters such as force, orientation, and surface contact. To address this, Pierrot~\emph{et al.}~\cite{pierrot1999hippocrate} implemented a force control schema in a teleoperative RUSS to maintain constant scanning force. Jiang~\emph{et al.}~\cite{jiang2020automatic} introduced an orientation optimization algorithm based on contact force for better visualization.
To ensure that the ultrasound probe is in good contact with the surface, Chatelain~\emph{et al.}~\cite{chatelain2017confidence} integrated ultrasound confidence map~\cite{karamalis2012ultrasound} into the control loop and presented confidence-driven control. However, the proposed control algorithm was designed for convex probe. In the sight of such limitation, Jiang~\emph{et al.}~\cite{jiang2022precise} proposed an orientation correction method based on the confidence map for the linear probes. The experimental results clearly demonstrated its effectiveness, but the theoretical justification is missing.

\par
Recent advancements in RUSS for vascular applications have gained significant attention. Jiang~\emph{et al.}~\cite{jiang2021autonomous} proposed an autonomous scanning framework for peripheral vascular diseases using real-time vessel segmentation. Bi~\emph{et al.}~\cite{bi2022vesnet} applied reinforcement learning to autonomously navigate the probe to the longitudinal view of the carotid artery. Huang~\emph{et al.}~\cite{huangQ2024robot} developed a system to autonomously perform carotid scans by imitating clinical protocols, while Goel~\emph{et al.}~\cite{goel2022autonomous} introduced a Bayesian Optimization-based path planning framework for femoral artery screening.

\par
\revision{
Most existing RUSSs for vascular scanning do not address the challenge of handling multiple vessels and often require additional tools to capture the doctor's intention for automatic maneuvering. Guidance can be implemented teleoperatively, with an expert operating remotely while the robot on-site follows~\cite{fu2022robot}, or through virtual reality, providing virtual guidance to the human operator~\cite{black2024human}. Another approach involves pre-planned scanning paths, where the robot provides virtual fixtures to enable reproducible ultrasound scanning for follow-up validations~\cite{huang2024robot}. However, among these strategies, gaze signal—an intuitive and hands-free interaction method—remains unexplored. This approach is particularly advantageous in intra-operative scenarios, as it allows surgeons to perform tasks without altering workflow.}

\subsection{Gaze-guided Medical Robotic Systems}
Initial attempts to integrate gaze information into medical robotic systems have been made in various scenarios, especially for laparoscopic surgeries. 
Noonan~\emph{et al.}~\cite{noonan2010gaze} used eye tracking to control an articulated robotic laparoscope for stable, hands-free visualization. Fujii~\emph{et al.}~\cite{fujii2018gaze} enhanced the gaze-guided laparoscope maneuverability by using a robotic arm instead of a rigid fixture. Clancy~\emph{et al.}~\cite{clancy2011gaze} combined gaze tracking with a liquid lens in the da Vinci system for automatic focus adjustment during minimally invasive surgeries. Gaze tracking has also been applied to constrain laparoscopic surgical tools for tissue safety~\cite{mylonas2012gaze} and to improve collaboration in multi-robot surgeries by visualizing fixation points~\cite{kwok2012collaborative}. Li~\emph{et al.}~\cite{li2018free} further optimized gaze tracking accuracy by compensating for head movements.
Beyond laparoscopic applications, Guo~\emph{et al.}~\cite{guo2019novel} utilized gaze to control needle insertion in CT-guided interventions, while Kogkas~\emph{et al.}~\cite{kogkas2019free} introduced a gaze-guided robotic scrub nurse to deliver surgical tools. To the best of our knowledge, initial attempt to integrate the gaze tracker into RUSS has not yet occurred.

\subsection{Gaze-guided Medical Image Analysis}\label{sec_gaze_med_IA}
Unlike optical images, medical images are often challenging to interpret, requiring solid biological knowledge for accurate diagnostics. To enhance the robustness of medical image analysis networks, human experts' gaze signals are frequently integrated as guidance. Cai~\emph{et al.}~\cite{cai2020spatio} proposed Temporal SonoEyeNet, which predicts sonographers' visual attention and ultrasound standard planes, demonstrating that understanding visual attention complements standard plane detection. Similarly, Wang~\emph{et al.}~\cite{wang2022follow} used class activation maps~\cite{zhou2016learning} to align network attention with human gaze.
In another application, Men~\emph{et al.}~\cite{men2023gaze} leveraged gaze tracking to improve probe movement prediction for obstetric standard plane navigation, while Alsharid~\emph{et al.}~\cite{alsharid2022gaze} showed its benefits in ultrasound video captioning. Most of these works aim to align networks with human attention to enhance reasoning. Beyond implicit guidance, gaze signals also serve as an intuitive tool for human-machine interaction. For instance, Khosravan~\emph{et al.}~\cite{khosravan2019collaborative} developed a collaborative lesion segmentation system that analyzes and filters doctors' gaze data to improve segmentation and reduce false positives.


\section{Methods}
As shown in Fig.~\ref{Fig_overview}, the proposed gaze-guided RUSS consists of two main components: image processing and robotic control. The human operator’s gaze signal is fed into an attention-estimation module, which uses sequential segmentation results and gaze tracking history to predict the operator's actual attention. It is important to distinguish between the gaze heatmap, directly captured by the eye tracker, and the attention heatmap, which reflects the operator's true focus.
Sonographers often glance at other areas of the ultrasound image to maintain contextual awareness of surrounding anatomical structures, rather than focusing solely on the vessel of interest. Differentiating these incidental glances from an intentional focus shift is critical to ensure the RUSS follows the correct vessel and avoids unnecessary target switches.
The estimated attention heatmap guides the gaze-guided segmentation network to extract the vessel of interest from the ultrasound image. The resulting segmentation directs the robot to center the vessel in the ultrasound image. To maintain good probe contact with the scanning surface, a confidence-driven orientation correction is implemented, adjusting the probe’s rotation around its y-axis based on the ultrasound confidence map to enhance image quality. During acquisition, impedance control~\cite{jiang2021autonomous} is employed to maintain consistent contact force.

\begin{figure*}[ht!]
\centering
\includegraphics[width=0.8\textwidth]{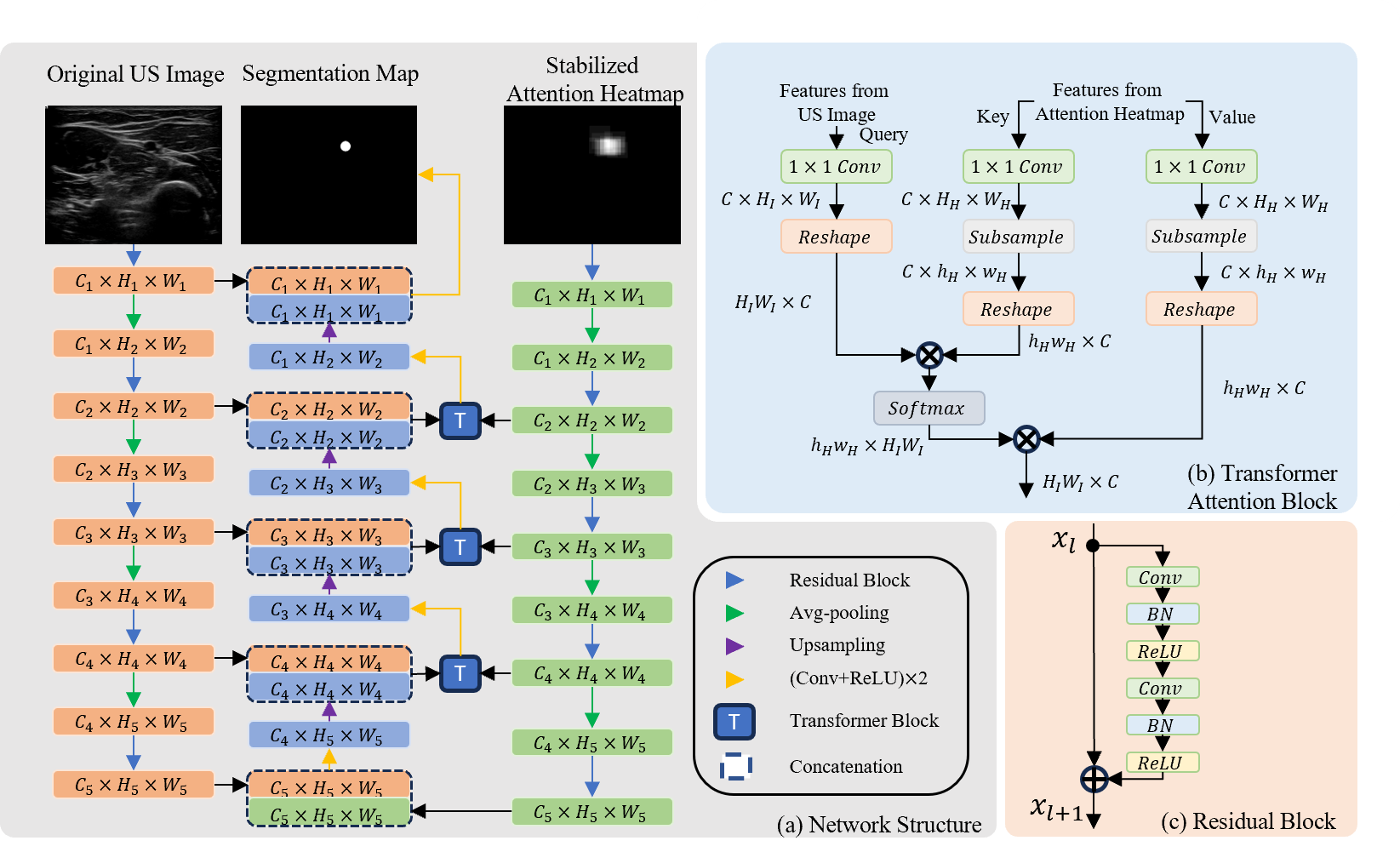}
\caption{(a) The overall design of the proposed gaze-guided segmentation network. (b) The structure of the transformer attention block. (c) The structure of the residual block.
}
\label{Fig_segmentation}
\end{figure*}

\subsection{Gaze-Guided Segmentation Network}\label{sec_gaze_guided_seg}
\revision{During the training of the proposed segmentation network, pseudo attention maps are generated from the label maps of B-mode ultrasound images and paired with ultrasound images to simulate human attention on vessels. This approach addresses the instability and noise in real gaze signals, which could mislead the network, and the difficulty of directly obtaining human attention heatmaps. By incorporating randomness into their generation, these pseudo attention maps create a more diverse and extensive dataset, reducing bias and improving the network's robustness and generalization.}
The pseudo attention heatmaps are generated based on the ground truth label maps. The centroid point of the heatmap ($X_m,Y_m$) is determined by the center point of the vessel label ($X_l,Y_l$).
\begin{align}
    (X_m, Y_m) &\sim \mathcal{N}\left( (X_l, Y_l), \Sigma_c \right) \\
    (x_m, y_m) &\sim \mathcal{N}\left( (X_m, Y_m), \Sigma_m \right)
\end{align}
where $\Sigma_c$ represents the covariance matrix of the Gaussian distribution. Based on the sampled centroid point of heatmap ($X_m,Y_m$), $N$ points ($x_m,y_m$) are selected and set to one.
The image that contains all the sampled points ($x_m,y_m$) is convolved by a $30\times30$ kernel with all weights setting to $1$. After the convolution, the resulting image is normalised to $[0,1]$. In this manner, a diffused heatmap $H$ is generated, simulating the density of the eye tracking points. Here $\Sigma_c\in R^{2\times2}$ and $\Sigma_m\in R^{2\times2}$ are diagonal matrices, with their diagonal values setting to 15 and 25, respectively.

\par
\revision{As depicted in Fig.~\ref{Fig_segmentation} (a), the segmentation network follows a U-shaped design. To enhance its performance, we incorporate attention heatmaps generated by the Human Intention Estimation Module based on human eye gaze signals. These maps represent the stabilized focus of the operator and are used to dynamically guide the network to prioritize the vessel of interest.}
\revision{A transformer attention block~\cite{vaswani2017attention} is incorporated to combine external attention signals from heatmaps with features extracted from ultrasound images. This adaptation enables the network to dynamically align its focus with the operator's intentions. As depicted in Fig.~\ref{Fig_segmentation} (b), the transformer attention block uses attention heatmaps as guidance to emphasize regions of interest. Inspired by~\cite{gao2021utnet}, subsampling is applied within the transformer block to reduce computational complexity. The transformed features are then concatenated with those from the ultrasound image encoder, facilitating the decoder's ability to generate precise segmentation masks.}

\par
If an ultrasound image contains multiple vessels, they are individually labeled and treated as distinct training samples to generate pseudo attention heatmaps. To reduce reliance on explicit guidance, $n\%$ of the generated attention maps are replaced with uniform maps where all weights are set to zero. This means that for $n\%$ of the training samples, no attention is provided, prompting the network to segment all vessels in the image independently. For this study, $n$ is set to $10\%$. This approach enhances the network’s adaptability, allowing gaze tracking to act as a performance booster rather than a critical component of the segmentation framework. During inference, pseudo attention heatmaps are replaced by the output of the attention estimation module (see Sec.~\ref{sec_estimation_m}).

\begin{figure}[ht!]
\centering
\includegraphics[width=0.4\textwidth]{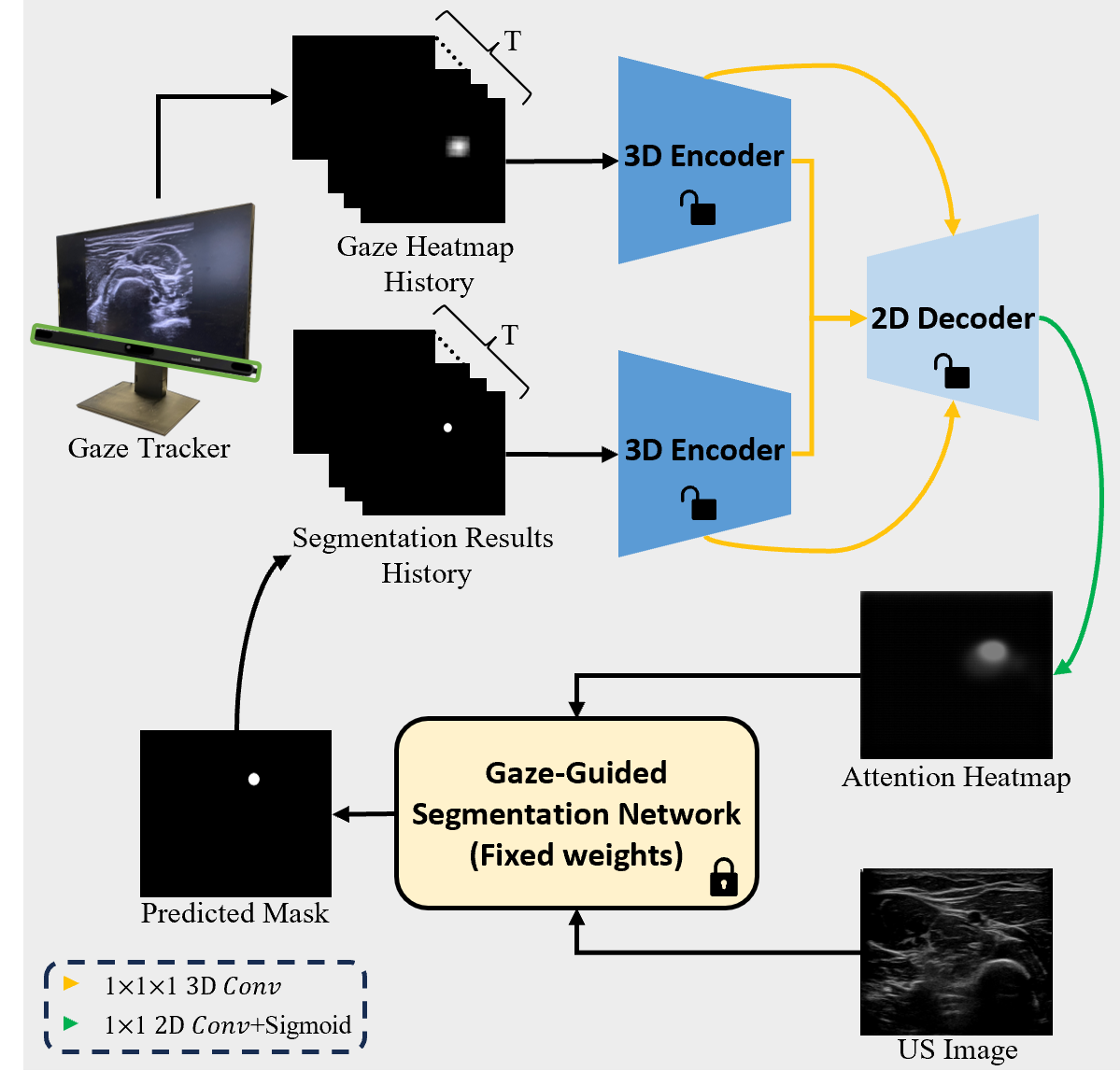}
\caption{The design of the human intention estimation module.
}
\label{Fig_estimation_module}
\end{figure}

\subsection{\revision{Stabilized Attention Heatmap Generation}}\label{sec_estimation_m}
Gaze heatmap is different from the human intention heatmap.
Although the attention of the operator remains focused on the target vessel, it is unrealistic for them to keep steering at it throughout the whole scanning process.
Instead, the sonographers tend to take glances at the surrounding anatomical structures to gain an overall contextual awareness. Therefore, without additional processing, the raw gaze heatmaps are not suitable to be directly used as attentions in the proposed segmentation network. In essence, the desired heatmap of the segmentation network should reveal the preference of the human operator among multiple vessels in the ultrasound image. 
\revision{To achieve this, the Human Intention Estimation Module analyzes sequential gaze patterns and segmentation label maps from previous timesteps, as shown in Fig.~\ref{Fig_estimation_module}. By combining these inputs, the module addresses the limitations of relying on gaze or segmentation history alone. For instance, if the segmentation history consistently focuses on a single vessel and the gaze heatmap briefly deviates, it is likely a temporal distraction. Conversely, stable deviation in the gaze heatmap suggests an intentional attention switch. This integration allows the network to reliably distinguish between temporal distractions and intentional changes in focus, ensuring robust guidance for segmentation.}

\par
\revision{As shown in Fig.~\ref{Fig_estimation_module}, in our implementation, T=64 (approximately 2.13 seconds) was chosen empirically. This value aligns well with the average human fixation duration ($290\pm155$ ms)~\cite{henderson2015neural} and avoids the time delay associated with higher T values.}
Two 3D convolutional encoders are employed to extract features from the gaze signals and segmentation results, respectively. At each scale, the extracted 3D feature maps are transformed into 2D feature maps through $1\times1\times1$ 3D convolution. The transformed 2D features are then concatenated with upsampled features in decoder to generate the final attention heatmap for the segmentation network.

\par
For training, different gaze patterns are recorded from human operators serving as training data. During training, the well-trained weights of the segmentation network is frozen and the output attention map from the attention estimation module is fed into the segmentation network (see Fig.~\ref{Fig_segmentation}) to perform segmentation task. 
\revision{The resulting DICE loss between the predicted segmentation map and the ground truth label was backpropagated to update the weights of the attention estimation module. By utilizing the fixed segmentation network, the DICE loss guided the attention estimation module to output refined attention maps that align with the segmentation task, effectively training the module to infer human attention.}

\begin{figure}[ht!]
\centering
\includegraphics[width=0.453\textwidth]{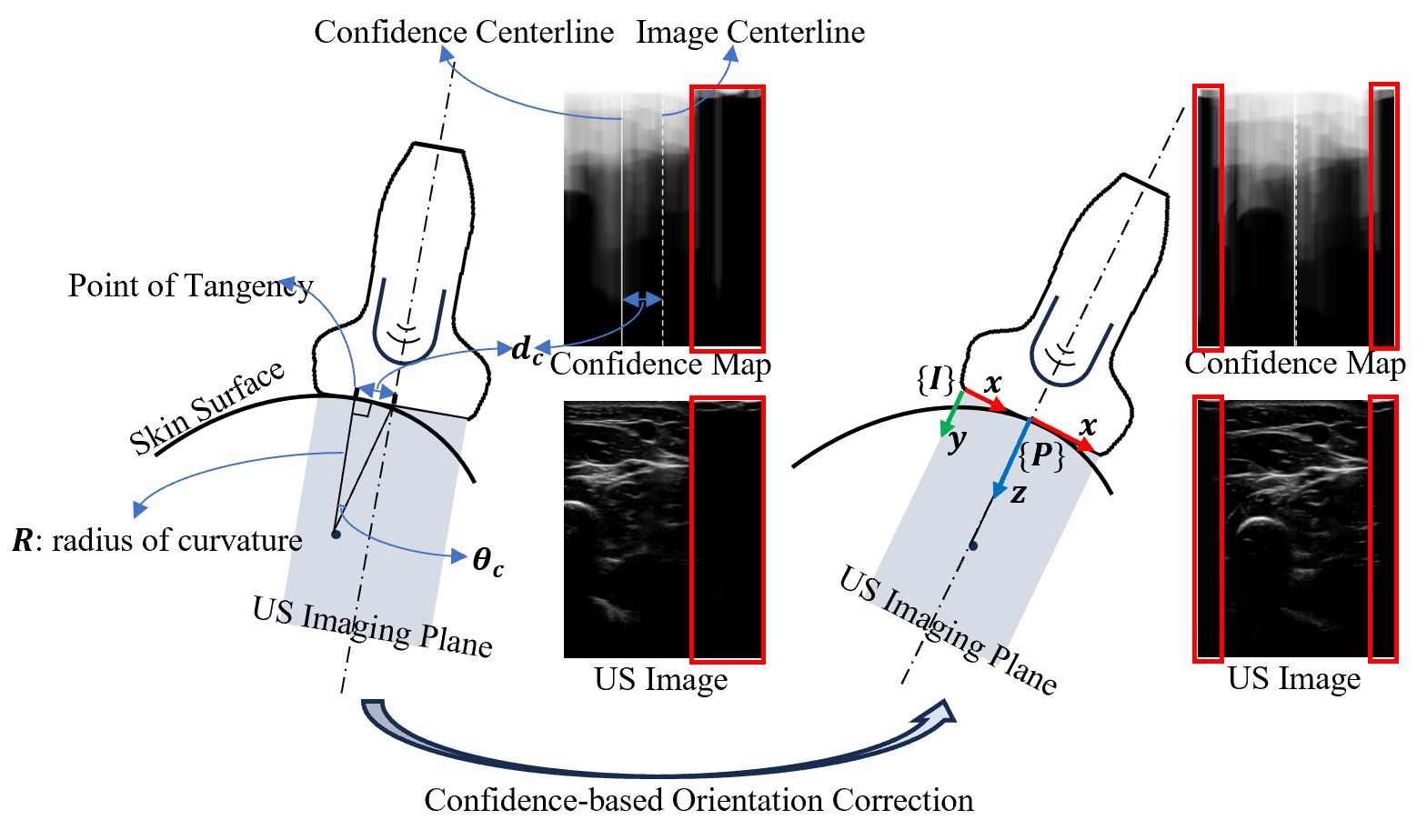}
\caption{The illustration of confidence-driven probe orientation correction for linear ultrasound probe. The red boxes in the images indicate the shadow areas caused by improper contact. $\{I\}$ represents the ultrasound imaging coordinate system, while $\{P\}$ is the probe coordinate system.
}
\label{Fig_conf_control}
\end{figure}

\subsection{Confidence-based Orientation Correction for Linear Probe}
The ultrasound probe's coordinate system is shown in Fig.~\ref{Fig_overview}. Translational movement along the z-axis is controlled using impedance control to ensure proper contact force with the scanning surface~\cite{jiang2021autonomous}. The y-axis represents forward motion during vessel scanning, while x-axis translation is guided by segmentation results to center the target vessel in the ultrasound image. In order to optimize the ultrasound image quality, a confidence-based orientation correction method for linear ultrasound probe is proposed.

\par
The confidence map~\cite{karamalis2012ultrasound} is widely used for assessing ultrasound image quality, assigning per-pixel confidence values based on a simplified sound propagation model. While the original confidence map relies on a computationally intensive random walk algorithm, the scan-line-based confidence map~\cite{chatelain2016quality} offers significantly improved efficiency.
To ensure real-time performance in the proposed confidence-driven control for the linear ultrasound probe, the scan-line-based confidence map is utilized. Let $I(x,y)$ denote the intensity value of a ultrasound image at position $(x,y)$ in the image coordination system. The scan-line-based confidence map is then given by:
\begin{equation}
    C(X,Y)=1-\left(\int_{0}^{{Y}_{max}}f(I(X,y))dy\right)^{-1}\int_{0}^{Y}f(I(X,y))dy
\end{equation}
where $f(*)$ is a non-linear mapping function between the radiofrequency signal intensity and the gray scale image intensity $I$. Here, $f(x) = x^{2}$ is used.

\par
As shown in Fig.~\ref{Fig_conf_control}, dark regions in the confidence maps (highlighted by red rectangles) indicate improper contact between the ultrasound probe and the skin surface. To enhance image quality, a correction angle ($\theta_c$) is calculated to rotate the probe around its y-axis, improving the confidence level in the central region of the ultrasound image. The confidence-weighted centerline is determined as follows:
\begin{equation}
    x_c = \frac{1}{C_\Omega}\iint_{(x,y)\in\Omega}xC(x,y)ydydx
\end{equation}
where $C_{\Omega} = \iint_{(x,y)\in\Omega}C(x,y)ydydx$.
\par
The position of the confidence-weighted centerline can be viewed as the tangency point between the linear transducer and the curved surface. Then as shown in Fig.~\ref{Fig_conf_control}, the distance ($d_c$) between the confidence-weighted centerline and the image centerline is computed.
\begin{equation}\label{eq_d_c}
    d_c = x_c-X_{c}
\end{equation}
where $X_c$ is the position of the image centerline. Then, the correction angle $\theta_c$ can be computed using Eq~(\ref{eq_conf_correction}):
\begin{equation}\label{eq_conf_correction}
    \theta_c = arctan(\frac{d_c}{R})
\end{equation}
where $R$ is the radius of curvature in the ideal case, this would allow for a one-time adjustment. However, due to the varying curvature of the human body, accurately estimating this parameter online for human tissue is challenging, particularly considering the pressure-induced deformation. To address this, angular correction was performed using visual servoing mode. The parameter $d_c$ ensures final convergence, aligning the confidence-based centerline with the centerline of the ultrasound imaging plane of the linear probe. A small $R$ may cause fluctuations in movement. To balance motion smoothness with angular correction efficiency, $R$ was empirically set to $10~cm$ based on experimental results.

\begin{table*}[ht!]
\caption{\revision{Segmentation results (Dice Score) of different segmentation network architectures and the meta data of each volunteer.}}\label{tab:segmentation}\centering

  \renewcommand{\arraystretch}{0.6}
  \resizebox{0.8\textwidth}{!}{
  \begin{tabular}{l l c| c| c| c| c}
    \toprule
    Method & Artery &$V_1$ & $V_2$ & $V_3$ & $V_4$ & $V_5$\\
    
    \midrule
    \multirow{2}{*}{UNet~\cite{ronneberger2015u}} & Radial & $0.456\pm 0.400$ & $0.562\pm 0.294$ & $0.084\pm 0.158$ & $0.401\pm 0.301$ & $0.292\pm 0.373$\\
    & Interosseous & $0.359\pm 0.276$ & $0.470\pm 0.297$ & $0.1821\pm 0.256$ & $0.467\pm 0.322$ & $0.389\pm 0.323$\\
    \cmidrule{2-7}
    \multirow{2}{*}{Att-UNet~\cite{schlemper2019attention}} & Radial & $0.486\pm 0.331$ & $0.611\pm 0.283$ & $0.254\pm 0.298$ & $0.708\pm 0.181$ & $0.385\pm 0.397$\\
    & Interosseous & $0.436\pm 0.283$ & $0.553\pm 0.224$ & $0.515\pm 0.377$ & $0.481\pm 0.340$ & $0.498\pm 0.278$\\
    \cmidrule{2-7}
    \multirow{2}{*}{UTNet~\cite{gao2021utnet}} & Radial & $0.671\pm 0.255$ & $0.710\pm 0.191$ & $0.275\pm 0.249$ & $0.748\pm 0.080$ & $0.263\pm 0.329$\\
    & Interosseous & $0.338\pm 0.207$ & $0.345\pm 0.314$ & $0.485\pm 0.288$ & $0.367\pm 0.250$ & $0.409\pm 0.302$\\
    \cmidrule{2-7}
    \multirow{2}{*}{UNet++~\cite{zhou2019unet++}} & Radial & $0.346\pm 0.332$ & $0.707\pm 0.221$ & $0.215\pm 0.223$ & $0.739\pm 0.090$ & $0.495\pm 0.259$\\
    & Interosseous & $0.442\pm 0.251$ & $0.421\pm 0.252$ & $0.581\pm 0.316$ & $0.242\pm 0.239$ & $0.549\pm 0.261$\\
    \midrule
    Ours + Pseudo  & Radial & $\mathbf{0.794\pm 0.130}$ & $\mathbf{0.740\pm 0.199}$ & $\mathbf{0.389\pm 0.346}$ & $\mathbf{0.880\pm 0.053}$ & $0.709\pm 0.263$\\
    Att. Heatmap & Interosseous & $0.603\pm 0.223$ & $0.672\pm 0.185$ & $0.741\pm 0.241$ & $\mathbf{0.755\pm 0.165}$ & $\mathbf{0.746\pm 0.166}$\\
    \cmidrule{2-7}
    Ours + Raw & Radial & $0.500\pm 0.423$ & $0.585\pm 0.351$ & $0.259\pm 0.309$ & $0.602\pm 0.384$ & $0.404\pm 0.336$\\
    Gaze Heatmap & Interosseous & $0.212\pm 0.235$ & $0.659\pm 0.167$ & $0.563\pm 0.290$ & $0.558\pm 0.327$ & $0.428\pm 0.329$\\
    \cmidrule{2-7}
    Ours + Est. & Radial & $0.736\pm 0.295$ & $0.721\pm 0.255$ & $0.335\pm 0.322$ & $0.822\pm 0.090$ & $\mathbf{0.737\pm 0.145}$\\
    Att. Heatmap & Interosseous & $\mathbf{0.642\pm 0.096}$ & $\mathbf{0.684\pm 0.125}$ & $\mathbf{0.764\pm 0.105}$ & $0.741\pm 0.156$ & $0.736\pm 0.106$\\
    \midrule
    \multicolumn{7}{c}{\revision{\textbf{Meta Data}}}\\
    \midrule
    \revision{BMI} &  &\revision{$26.6$} & \revision{$24.9$} & \revision{$24.5$} & \revision{$26.9$} & \revision{$21.4$}\\
    \cmidrule{2-7}
    \multirow{2}{*}{\revision{Average Size}} & \revision{Radial} &\revision{$3.0~mm$} & \revision{$2.3~mm$} & \revision{$1.7~mm$} & \revision{$2.8~mm$} & \revision{$2.6~mm$}\\
     & \revision{Interosseous} &\revision{$3.3~mm$} & \revision{$2.5~mm$} & \revision{$2.8~mm$} & \revision{$3.1~mm$} & \revision{$2.8~mm$}\\
    \bottomrule
  \end{tabular}
  }
\end{table*}

\section{\final{Experiments and Results}}

\subsection{Experimental Setup}\label{sec:implementation}
To validate the proposed gaze-guided segmentation framework, ultrasound images of radial and interosseous arteries were collected using the Siemens Juniper ultrasound machine (ACUSON Juniper, SIEMENS AG) under Institutional Review Board approval from the Technical University of Munich (reference number 2022-87-S-KK). Ten ultrasound sweeps were obtained from five adult male volunteers, covering both arms, resulting in $2421$ images. Image labeling was conducted under clinical supervision using ImFusionSuite (ImFusion GmbH).
While the dataset size is relatively small, this reflects the practical challenges of collecting labeled data in medical imaging. The primary aim of this study is to validate how gaze signals can facilitate autonomous vessel segmentation, even with limited data availability, a common issue in medical applications.
To train the attention estimation module, gaze data was collected from two volunteers instructed to focus on either the radial or interosseous artery during ultrasound sweeps. Each volunteer concentrated on a single vessel while the sweep video played, capturing their ground truth intention. To enhance robustness, volunteers were allowed to occasionally glanced at other image regions to reflect natural gaze variability. A total of 160 gaze recordings were obtained using the Tobii Eye Tracker 5, with 16 recordings per sweep.
The networks were trained using the Adam optimizer with a fixed learning rate of $1\times10^{-5}$ on a single GPU (Nvidia GeForce RTX 4070).

\par
The proposed gaze-guided robotic ultrasound scanning system comprises a robot arm (KUKA LBR iiwa 7 R800, KUKA Roboter GmbH), a linear ultrasound probe (12L3, Siemens AG) attached to the robot via a 3D-printed probe holder, and a gaze tracker (Tobii Eye Tracker 5, Tobii AB) mounted below a display screen. The gaze tracker is calibrated using the manufacturer-provided software (Tobii Experience, Tobii AB), and real-time gaze signals are visualized as heatmap on the screen for user feedback. Data collection focused exclusively on eye-focusing heatmaps of ultrasound images, ensuring anonymity by avoiding the capture of participants' faces or eyes. \final{Consent was obtained from all volunteers prior to their participation.}
The robot operates via a Robot Operating System (ROS) interface, and ultrasound images are accessed using a frame grabber (USB Capture HDMI Plus, Magewell Electronics). A commercial arm phantom (BPA304, Blue Phantom GmbH) is used for evaluation.

\subsection{Results of Gaze-guided Segmentation Network}\label{sec_segmentation_results}
The ultrasound dataset described in Sec.~\ref{sec:implementation} is used to validate the segmentation performance of the proposed network against other segmentation networks. For individual volunteer performance, five models are trained for each network structure, with four volunteers used for training and the remaining one for testing. The models are indexed as $V_1$ to $V_5$ in Table~\ref{tab:segmentation}.
Unlike other networks, which require separate weights for radial and interosseous arteries, the proposed network uses a single set of weights to segment both arteries. The target vessel is automatically determined based on gaze tracking results. During training, pseudo attention heatmaps are generated separately for the radial and interosseous artery labels. The Dice Score is used as the evaluation metric.
This section evaluates the performance of the gaze-guided segmentation network using pseudo attention heatmaps, while Sec.~\ref{sec_estimation_results} examines performance with estimated attention heatmaps.

\par
The segmentation results are shown in Table~\ref{tab:segmentation}. The proposed network outperforms all the SOTA segmentation networks.
We can observe that without the guidance of gaze signal, the segmentation of interosseous arteries can hardly achieve a dice score of $0.55$. Conversely, segmentation performance on radial arteries occasionally surpasses 0.70 under similar conditions.
Such phenomenon is due to the fact that interosseous artery is located deeper in the arm, while radial artery is more superficial to the skin surface. Based on the propagation principle of ultrasound waves, the deeper part of the ultrasound image usually have more severe attenuation effect than the upper part of the image. Therefore, the visibility and clearness of interosseous artery is in general worse than the radial artery. 
An exception is the case of $V_3$, for $V_3$, the lower segmentation performance across all methods can be attributed to the extraordinarily small size of the radial artery ($1.7$ mm), which is a significant outlier compared to the average adult male radial artery size ($2.68\pm0.24$~\cite{wahood2022radial}). The difficulty of segmenting such an extreme outlier highlights the limitations of the current dataset. In contrast, the significant improvement in $V_5$'s segmentation performance with gaze guidance is due to the thinner subcutaneous tissue layer resulting from a lower BMI as shown in Table~\ref{tab:segmentation}. The proposed gaze-guided method successfully leveraged the gaze signal to focus on the target region, despite structural differences in the data.
Besides, for the segmentation of radial artery, in some cases, UTNet and UNet++ perform almost equally well as our proposed method, especially in the case of $V_2$, where UTNet achieved slightly worse mean value with smaller variance. This is because the radial artery is generally easy to identify from the ultrasound images. 
\final{Moreover, we need to be aware that UTNet is trained to specifically focus on one artery at a time. Whereas the proposed gaze-guided segmentation network is trained to segment both vessels individually at a time.}
In contrast, for the segmentation of interosseous artery, the proposed gaze guided network clearly outperforms all the other network structures. The explicit guidance from human operator, in this case, can not only help the network to determine the segmentation target but also provide vital guidance to the network to locate challenging structures.

\subsection{Results with Attention Estimation Module}\label{sec_estimation_results}
In this section, we combine the attention estimation module with the segmentation module to validate their joint performance. 
\revision{To train the attention estimation module, 160 gaze recordings were captured (16 per ultrasound sweep). For half of the recordings, volunteers were instructed to focus on the radial artery, while for the other half, they were asked to focus on the interosseous artery. For each gaze recording, volunteers were instructed to focus on only one of the two vessels for the entire duration of the ultrasound video. Of these, 120 recordings were used for training and 40 (four per sweep, two per vessel) for testing.}
To prevent the network from overly relying on perfect segmentation results, imperfect segmentation masks were generated using unprocessed gaze heatmaps. Training samples included a mix of perfect labels, imperfect segmentation masks, and blank images. This approach avoids the estimation module relying solely on segmentation history and ensures proper utilization of gaze heatmaps. As noted in Sec.~\ref{sec_estimation_m}, during training, the gaze-guided segmentation network weights are fixed, and only the weights of the estimation module are updated.

\par
The lower half of Table~\ref{tab:segmentation} presents the segmentation results of the gaze-guided network with and without the attention estimation module. Results without the estimation module were obtained by directly feeding raw gaze heatmaps into the network, following the same testing strategy as Sec.~\ref{sec_segmentation_results}. The trained gaze-guided segmentation network from Sec.~\ref{sec_segmentation_results} was used to train the corresponding estimation module, and the matching test gaze recordings were employed for evaluation.
\final{The absence of the estimation module significantly hinders segmentation performance, largely due to the noise and instability of raw gaze signals, compounded by gaze tracker calibration errors. The segmentation results with the estimation module closely match those achieved using pseudo attention heatmaps.} In most cases, the pseudo attention heatmap delivers the best performance. However, in some instances, the estimated attention heatmap outperforms the pseudo attention heatmap, with differences within $0.04$, likely due to data imbalance. The execution time of segmentation and attention estimation modules is $36.1 \pm 0.5$ ms.

\subsection{Gaze-Guided Robotic Ultrasound Scanning}

\begin{figure}[ht!]
\centering
\includegraphics[width=0.4\textwidth]{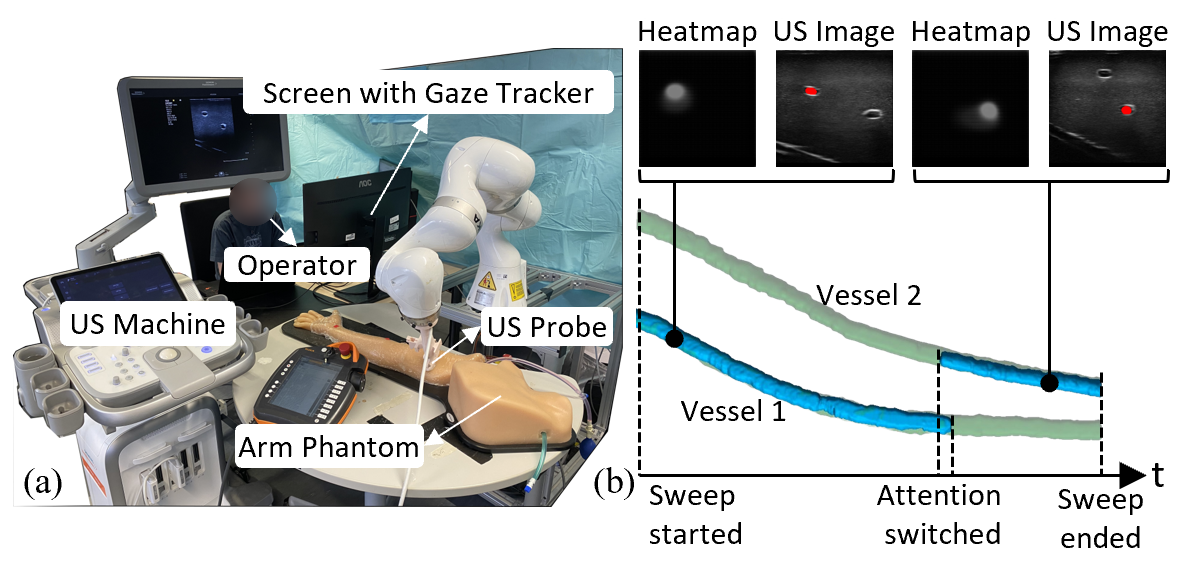}
\caption{(a) The experimental setup of the gaze-guided RUSS. (b) Reconstruction results of one representative scanning.
}
\label{Fig_exp_setup}
\end{figure}
As shown in Fig.~\ref{Fig_exp_setup}(a), experiments were conducted on an arm phantom to validate the proposed gaze-guided RUSS. The start and end points of the scans were manually selected, with an average scanning length of $12.9 \pm 0.7$ cm. An eye tracker mounted below the screen captured the operator's gaze signals. The segmentation network was trained on phantom data, while the attention estimation module was adapted from the model trained in Sec.~\ref{sec_estimation_results}.
Fig.~\ref{Fig_exp_setup}(b) displays the ultrasound reconstruction of the two vessels in the arm phantom. The blue regions represent reconstruction results from the proposed network’s segmentation masks, overlaid on the ground truth vessel model in green. Both vessels were segmented successfully as the operator’s attention switched from vessel 1 to vessel 2, demonstrating smooth and effective vessel switching.
\begin{table}[ht!]
\caption{Performances w/ and w/o confidence-based orientation correction (Averaged $d_c$).}\label{tab:confidence_control}\centering
  \renewcommand{\arraystretch}{0.6}
  \resizebox{0.45\textwidth}{!}{
  \begin{tabular}{L{0.08\textwidth} |C{0.16\textwidth} |C{0.16\textwidth}}
    \toprule
     & w/o Confidence-based Orientation Correction & w/ Confidence-based Orientation Correction\\
    \midrule
   Sweep 1 & $1.42\pm1.43$ mm & $7.21\pm0.77$ mm \\
   Sweep 2 & $1.70\pm2.03$ mm & $7.11\pm0.92$ mm \\
   Sweep 3 & $1.92\pm1.63$ mm & $8.95\pm3.18$ mm \\
    \bottomrule
  \end{tabular}
  }
\end{table}

\par
Experiments were also conducted to validate the control performance of the confidence-based orientation correction. We conducted $3$ sweeps without the confidence-based control and $3$ sweeps with the confidence-based control activated. Table~\ref{tab:confidence_control} shows the averaged $d_c$ from Eq.~\ref{eq_d_c}, which calculates how much the confidence centerline is deviated from the image centerline. A high value of $d_c$ indicates the existence of a big portion of shadow area in the image. The proposed confidence-based method can efficiently optimize the ultrasound confidence by adjusting the in-plane orientation of the ultrasound probe.

\section{Conclusion and Discussions}
In this work, a gaze-guided RUSS is proposed. In order to integrate the human gaze signal into the control loop of RUSS, we proposed a gaze-guided segmentation network. Since the human gaze signal is noisy and unstable, a human intention estimation module is proposed to reveal the essence human attention from the gaze signal. Experiments have been conducted on the ultrasound images of radial and interosseous arteries. The proposed network demonstrated superior segmentation performance compared to SOTA methods.
\final{We acknowledge the limited dataset size and plan to expand it with more diverse anatomical structures and patient data to enhance generalizability. Additionally, future work will address the generalization challenge by exploring novel network architecture designs~\cite{bi2023mi}. These efforts aim to further validate and improve the system’s adaptability.}
Finally, the whole system was tested on an arm phantom with robots. The robot is controlled to centralise the vessel of interest based on the operator's preference interpreted from the gaze signal. To make sure the robot can establish proper contact with the scanning surface, a confidence-based orientation adjustment method is proposed. The phantom test validated the proficiency of the whole system. In the future, we would further validate the proposed system on real human tissue.